\documentclass[10pt,twocolumn,letterpaper]{article}

\usepackage[pagenumbers]{cvpr} 

\usepackage[dvipsnames]{xcolor}
\newcommand{\red}[1]{{\color{red}#1}}

\usepackage{graphicx}
\usepackage{amsmath}
\usepackage{amssymb}
\usepackage{booktabs}

\definecolor{cvprblue}{rgb}{0.21,0.49,0.74}
\usepackage[pagebackref,breaklinks,colorlinks,citecolor=cvprblue]{hyperref}

\usepackage[capitalize]{cleveref}
\crefname{section}{Sec.}{Secs.}
\Crefname{section}{Section}{Sections}
\Crefname{table}{Table}{Tables}
\crefname{table}{Tab.}{Tabs.}

\def\paperID{10822} 
\def\confName{CVPR}
\def\confYear{2024}

\usepackage{color}
\usepackage{xcolor}
\definecolor{deepGreen}{RGB}{0,153,0}
\definecolor{orange}{RGB}{255,125,0}
\def\red#1{\textcolor[rgb]{1,0,0}{#1}}

\definecolor{sainone}{RGB}{236, 242, 249} 
\definecolor{saintwo}{RGB}{255, 230, 204}

\newcommand{\cut}[1]{}
\usepackage{multirow}
\usepackage{amsmath}
\usepackage{amssymb}
\usepackage{mathtools}
\usepackage{arydshln}
\usepackage{soul}
\usepackage{nicefrac}
\usepackage{booktabs}
\usepackage{stmaryrd}
\usepackage{caption}
\usepackage{graphicx}
\usepackage{colortbl}
\definecolor{gray}{gray}{0.9}
\definecolor{pink}{RGB}{255, 234, 232}
\sethlcolor{pink}
\usepackage[accsupp]{axessibility}
\usepackage{scalerel,graphicx,xparse}

\usepackage{array}
\usepackage{float}
\usepackage{arydshln}

\makeatletter
\apptocmd\@maketitle{{\myfigure{}\par}}{}{}
\makeatother
\usepackage{capt-of,etoolbox}

\makeatletter
\newcommand\notsotiny{\@setfontsize\notsotiny\@vipt\@viipt}
\makeatother

\usepackage{xcolor,pifont}
\newcommand*\colourcheck[1]{%
  \expandafter\newcommand\csname #1check\endcsname{\textcolor{#1}{\ding{52}}}%
}
\newcommand*\colourcross[1]{%
  \expandafter\newcommand\csname #1cross\endcsname{\textcolor{#1}{\ding{55}}}%
}
\colourcheck{blue}
\colourcheck{green}
\colourcheck{black}
\colourcross{red}
\colourcross{black}

\newcommand{\MYhref}[3][brown]{\href{#2}{\color{#1}{#3}}}%
\title{\vspace{-6mm}\textit{DemoCaricature}: Democratising Caricature Generation with a Rough Sketch \vspace{-0.6cm}}

\author{
\MYhref{https://github.com/ChenDarYen}{Dar-Yen Chen} \hspace{.2cm} 
\MYhref{https://ayankumarbhunia.github.io/}{Ayan Kumar Bhunia} \hspace{.2cm} 
\MYhref{https://subhadeepkoley.github.io/}{Subhadeep Koley} \hspace{.2cm}
\MYhref{https://aneeshan95.github.io/}{Aneeshan Sain} \\ 
\MYhref{https://www.pinakinathc.me/}{Pinaki Nath Chowdhury} \hspace{.2cm}
\MYhref{https://www.surrey.ac.uk/people/yi-zhe-song}{Yi-Zhe Song} \\
SketchX, CVSSP, University of Surrey, United Kingdom.  \\
{\tt\small \{d.chen, a.bhunia, s.koley, a.sain, p.chowdhury, y.song\}@surrey.ac.uk}\\
\small{\url{https://democaricature.github.io}}}
\vspace{-0.4cm}

\newcommand\myfigure{
\centering
\vspace{-0.7cm}
\captionsetup{type=figure} 
    \includegraphics[width=.94\textwidth]{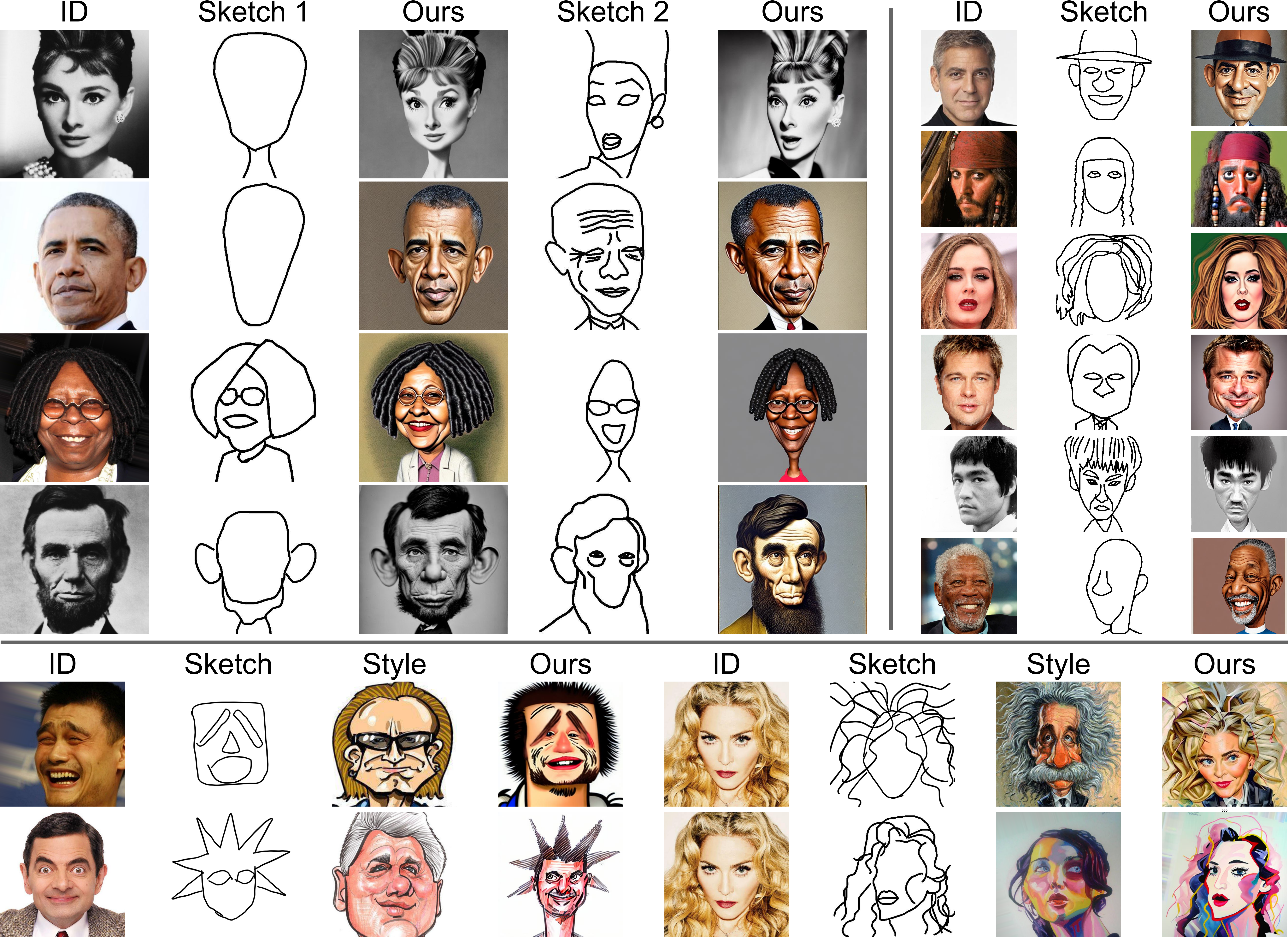}
    \vspace{-0.35cm}
\captionof{figure}{Given an \textit{abstract freehand} sketch and an image depicting the facial identity of a person, our method transforms the deformed sketch into a plausible-looking caricature while maintaining \textit{identity-fidelity} and imitating the \textit{exaggerations} portrayed in the input sketch. Additionally, it can seamlessly transmit the \textit{look-and-feel} of a given \textit{style-image} into the output caricature.}

\label{fig:teaser}
\vspace{+0.3cm}
}

\begin{document}
\maketitle
\begin{abstract}
\vspace{-0.4cm}

In this paper, we democratise caricature generation, empowering individuals to effortlessly craft personalised caricatures with just a photo and a conceptual sketch. Our objective is to strike a delicate balance between abstraction and identity, while preserving the creativity and subjectivity inherent in a sketch. To achieve this, we present Explicit Rank-1 Model Editing alongside single-image personalisation, selectively applying nuanced edits to cross-attention layers for a seamless merge of identity and style. Additionally, we propose Random Mask Reconstruction to enhance robustness, directing the model to focus on distinctive identity and style features. Crucially, our aim is not to replace artists but to eliminate accessibility barriers, allowing enthusiasts to engage in the artistry.


\end{abstract}

\vspace{-7mm}
\section{Introduction}
\vspace{-0.2cm}
\label{sec:intro}

Ever wondered when you would finally decide to get that personalised caricature created, perhaps during a holiday? Look no further, this paper is for you -- we strive to democratise caricature \cite{cao2018carigans, HuoBMVC2018WebCaricature, Jang2021StyleCari} generation for everyone! With a portrait of yourself and a conceptual sketch of how you envision your caricature, we will automatically generate a high-fidelity caricature that unmistakably captures your essence \cite{Jang2021StyleCari}. \cut{We emphasise that}Our aim however is not to replace artists; after all, the realm of art may be one that AI will never entirely conquer -- so when you find yourself in Paris next, \textit{do} get your caricature expertly crafted by a skilled artist!

We commence our study by asking the fundamental question that arises when scrutinising a caricature -- Is this me? (or Obama or Mr.\ Bean for that matter in \cref{fig:teaser}?) Indeed, the core challenge in caricature generation is navigating the delicate balance of infusing abstraction \cite{hertzmann2020line} into the process to achieve that distinctive caricature appearance, while still preserving the essential identity cues that unmistakably represent the intended person \cite{deng2019arcface}. Over and above all, how do we seamlessly inject your individuality \cite{chowdhury2023what} and creativity \cite{bhunia2022doodle} into the art generation process, ensuring the resulting caricature is genuinely \textit{your own}, rather than one dictated \textit{solely} by AI?

Our solution lies in your sketch! A single rough sketch \cite{koley2024how, bandyopadhyay2024doodle, bandyopadhyay2024xai} is all it takes to encapsulate \textit{your} vision for \textit{your} caricature, as illustrated in \cref{fig:teaser}. The scientific challenge is clear: regardless of your artistic skill or the lack of it, how can we design a system to adeptly generate a plausible caricature while still preserving your identity \cite{Jang2021StyleCari}? And one more thing, if there is a specific caricature style you prefer, we would like to accommodate that preference as well.

We most certainly are not pioneers in caricature generation~\cite{cao2018carigans, Jang2021StyleCari, HuoBMVC2018WebCaricature}; our motivation primarily draws from prior art in this field. However, our set of challenges notably surpasses the technical capabilities of previous systems \cite{Jang2021StyleCari, cao2018carigans}, particularly those primarily deformation-based \cite{shi2019warpgan, gu2021carime}, which tend to prioritise style creation over identity. Crucially, these prior systems often fall short in including ``you'' in the solution. This deficiency results in generated caricatures lacking expressiveness and missing interesting features like local abstraction \cite{hertzmann2020line}, hairstyle variation \cite{wei2022hairclip}, and view changes -- all of which can be easily injected into our system with just your single sketch!

Our approach to modelling the delicate balance between identity and style relies on the interaction between a novel single-image Text-to-Image (T2I) personalisation module and a sketch-specific T2I-Adapter~\cite{mou2023t2i}. The former ensures identity, while the latter allows for sketch-controlled caricature generation. This, of course, is not trivial. Latest single-image T2I personalisation approaches \cite{gal2023an, ruiz2022dreambooth, tewel2023keylocked} often grapple with overfitting during single-image fine-tuning, resulting in a highly specialised yet inflexible model that lacks generalisation beyond training data. This makes them especially challenging for them to adapt to the highly exaggerated and subjective human sketches, which are often Out-of-Distribution (OOD). This challenge is further exacerbated, as we face the task of merging concepts of identity and style. If done blindly, this would lead to a blending of features, resulting in caricatures that lack distinction or skew towards one aspect at expense of the other.

\cut{For this purpose} We thus propose Explicit Rank-1 Model Editing for single-sketch personalisation, enabling effective learning and the fusion of identity and style. By incorporating an explicit mechanism, it independently manipulates the explicit editing of identity and style in the cross-attention layers~\cite{mou2023t2i}, with minimal extra parameters while maintaining the integrity of textual contexts. This provides a more subjective and fine-grained control over desired concepts, mitigating the overfitting typically encountered. Furthermore, we introduce Random Mask Reconstruction to enhance the robustness of distorted shapes. It achieves this by masking random patches of the input image, compelling the model to focus on crucial identity and style features over local variations. This capability importantly allows the model to better handle exaggerated caricature sketches while emphasising the essential learned features.

Our contributions are: \textit{(i)} we democratise caricature generation, enabling individuals to easily create personalised caricatures, from a photo and a conceptual sketch. \textit{(ii)} we address the delicate balance between abstraction and identity via Explicit Rank-1 Editing, offering nuanced control by selectively applying rank-1 edits to cross-attention layers. \textit{(iii)} we enhance system robustness with Random Mask Reconstruction, enabling effective handling of distorted shapes while emphasising essential identity and style.

\vspace{-3mm}
 \section{Related Work}
\label{sec:related}
\vspace{-1mm}
\noindent \textbf{Deep Caricature Synthesis: }
Caricature synthesis aims to \textit{exaggerate} or \textit{distort} specific facial features for a \textit{stylised} yet \textit{recognisable} portrayal of a subject \cite{gong2020autotoon,gu2021carime,Bazazian_2022_CVPR}. Such methods typically involve a deformation stage, followed by image-to-image translation. Introduced as a GAN~\cite{goodfellow2014generative}-based framework \cite{cao2018carigans} involving facial landmarks to guide deformations, it was enhanced by automating control point prediction for warping and embedding a discriminator, acting as an \textit{identity} classifier to help in its preservation \cite{shi2019warpgan}. A few subsequent works include diversifying caricature generation to multiple facial exaggeration types \cite{gu2021carime}, leveraging SENet \cite{hu2018senet} and spatial transformer modules to produce high-fidelity warps based on dense warping field \cite{gong2020autotoon}, and leveraging StyleGAN \cite{karras2019style} with GAN inversion \cite{Piao_2021_CVPR, tewari2020pie} to propose \textit{shape exaggeration} blocks for additional control \cite{Jang2021StyleCari}. Towards spatial manipulation within caricature synthesis, while Semantics CariGAN \cite{chu2021learning} leveraged semantic shape transformations for caricature-control from warped semantic maps, a segmentation-guided dual-domain synthesis framework \cite{Bazazian_2022_CVPR} combined few-shot GAN \cite{robb2020fsgan} with RepurposingGAN \cite{Tritrong2021RepurposeGANs}. Addressing the limitations of the deformation-based pipeline, we strive to enhance creative freedom in caricature synthesis via freehand sketches.




\vspace{+1mm}
\noindent \textbf{Denoising Diffusion Probabilistic Models (DDPM):} Recently, DDPMs \cite{ho2020denoising} have emerged as the de facto choice for generative modelling, thanks to their high-fidelity image synthesis potential \cite{koley2024text}. Earlier works \cite{ho2020denoising, ho2021classifierfree, pmlr-v162-nichol22a, rombach2022high} have significantly improved text-to-image (T2I) models, such as Imagen \cite{saharia2022photorealistic}, DALL-E2 \cite{ramesh2022hierarchical}, and Stable Diffusion (SD) \cite{rombach2022high} -- further enhanced by training on diverse image-caption pair datasets \cite{schuhmann2021laion400m, schuhmann2022laionb}. Harnessing the prior knowledge of \textit{pretrained} T2I models, research progressed to guide generation under additional conditions \cite{zhao2023uni, ye2023ip-adapter, xu2023prompt}. For instance, ControlNet \cite{Zhang_2023_ICCV} and T2I-Adapter \cite{mou2023t2i} introduced content semantics adapters for targeted tasks such as pose, depth map, and sketch-conditional synthesis \cite{wang2022pretraining}, which enhances the flexibility of the generation process.

\noindent \textbf{T2I Personalisation:} With a limited set of reference images, T2I personalisation aims to adapt pretrained T2I models \cite{saharia2022photorealistic, rombach2022high} to specific concepts, while retaining its generalisability. Among the proposed strategies \cite{voynov2023P+, sohn2023styledrop, dong2023dreamartist}, while Textual Inversion \cite{gal2023an} optimises text embeddings to capture new concepts, DreamBooth \cite{ruiz2022dreambooth} personalises the output by fine-tuning the whole Stable Diffusion \cite{rombach2022high} and Imagen \cite{saharia2022photorealistic} models. Research on Parameter-Efficient Fine-Tuning (PEFT) methods \cite{he2023sensitivity}, such as LoRA \cite{hu2022lora, db_lora} and SVDiff \cite{Han_2023_ICCV}, focuses on reducing the computational burden during model training. Additionally, CustomDiffusion \cite{kumari2023multi} fine-tunes only cross-attention layers, while Perfusion \cite{tewel2023keylocked} introduces Rank-1 Model Editing (ROME) \cite{meng2022locating} to optimise the Value-pathway in the cross-attention mechanism. InstantBooth \cite{shi2023instantbooth} enables personalised inference with single images. FastComposer \cite{xiao2023fastcomposer} uses a novel image encoder for concept embeddings, while HyperDreamBooth \cite{ruiz2023hyperdreambooth} achieves efficient fine-tuning with a hypernetwork. However, resource-intensive training may limit their application \cite{shi2023instantbooth, xiao2023fastcomposer}. We thus offer a rapid and universal single-image method, that extends personalisation beyond individual identity images to include reference style images, facilitating synthesis of stylised caricatures, while costing minimal iterations and parameter overhead.


\vspace{-2mm}
\section{Revisiting Text-to-Image Diffusion Models}
\vspace{-1mm}
\noindent \textbf{Overview:}
Diffusion models~\cite{rombach2022high, dhariwal2021diffusion, tang2023emergent}, rely on two stochastic processes, termed as \textit{forward} and \textit{backward} diffusion~\cite{ho2020denoising}. The \textit{forward process} involves iteratively adding Gaussian noise to a clean image ${x}_0 \in \mathbb{R}^{h\times w\times 3}$ over $t$ time-steps, producing a noisy image ${x}_t \in \mathbb{R}^{h\times w\times 3}$ as:
%
%
$
{x}_t = \sqrt{\bar{\alpha}_t}~{x}_{0} + (\sqrt{1-\bar{\alpha}_t})\epsilon
$,
where, $\epsilon\sim\mathcal{N}(0,\mathbf{I})$ is the added noise, $\{\alpha_t\}_1^T$ represents a predefined noise schedule~\cite{ho2020denoising} with $\alpha_t=\prod_{s=1}^{T}\alpha_s$ and time-step $t$ value is sampled from a Uniform distribution $t\sim U(0,T)$. With sufficiently large $T$, $x_T$ approximates isotropic Gaussian noise. The \textit{backward process} entails training a modified UNet~\cite{ronneberger2015u} denoiser $F_\theta(\cdot)$. It takes the noisy input ${x}_t$ and the corresponding time-step $t$ to estimate the input noise $\epsilon_t \approx F_\theta({x}_t,t)$. Once trained with a standard MSE loss~\cite{ho2020denoising}, $F_\theta$ can reverse the effect of forward diffusion. During inference, $F_\theta$ is applied iteratively for $T$ time-steps on a randomly sampled 2D Gaussian noise image ${x}_T$ to get a cleaner image ${x}_{t-1}$ at each time-step $t$, thus eventually resulting in  one of the cleanest images ${x}_0$ resembling the original target distribution~\cite{ho2020denoising}


\vspace{0.1cm}
\noindent \textbf{Text-Conditioned Diffusion Model:}
Diffusion models can generate images conditioned on different signals (\textit{e.g.}, class labels~\cite{ho2022cascaded}, textual prompts~\cite{rombach2022high, ramesh2022hierarchical}, etc.). 
Given a textual prompt $p$, the initial step involves its conversion to the word-embedding space $\mathcal{W}$ on applying a word-embedding function $\mathbf{W}$. Subsequently, the transformed prompt is passed through a CLIP~\cite{radford2021learning} text encoder denoted by $\mathbf{T}(\cdot)$, which produces the text encoding as $\mathbf{t_p} = \mathbf{T}(\mathbf{W}(p)) \in \mathbb{R}^{77 \times d}$ in the text encoding space $\mathcal{T}$. This $\mathbf{t_p}$ controls the diffusion process via cross-attention, thus allowing $F_\theta(x_t,t,\mathbf{t_p})$ to perform $p$ controlled  denoising on $x_t$.





\vspace{0.1cm}
\noindent \textbf{Stable Diffusion:}
Latent Diffusion Models (\ie, Stable Diffusion)~\cite{rombach2022high} perform forward and backward denoising in the \textit{latent} space for \cite{rombach2022high}. In its \textit{two-stage} approach, Stable Diffusion (SD)~\cite{rombach2022high} first trains a \textit{variational autoencoder} (VAE) \cite{kingma2013auto}, comprising an encoder $E(\cdot)$ and a decoder $D(\cdot)$ in sequence. $E(\cdot)$ converts the input image $x_0\in\mathbb{R}^{h\times w\times c}$ to its latent representation $z_0\in\mathbb{R}^{{\frac{h}{8}}\times {\frac{w}{8}}\times d}$ \cite{rombach2022high}. The {forward process} adds Gaussian noise to ${z}_0$ over $t$ time-steps, producing a noisy latent ${z}_t = \sqrt{\bar{\alpha}_t}~{z}_{0} + (\sqrt{1-\bar{\alpha}_t})\epsilon$. Later, a UNet \cite{ronneberger2015u} denoiser $\epsilon_\theta(\cdot)$ is trained to perform conditional denoising based on textual prompt $p$ directly in the latent space~\cite{rombach2022high} with loss objective as:
\vspace{-0.3cm}
\begin{equation}
    \mathcal{L}_{sd} = \mathbb{E}_{z_t,t,\epsilon,p}(||\epsilon-\epsilon_\theta(z_t,t,\mathbf{t_p})||_2^2)
    \vspace{-0.2cm}
    \label{eq:ldmloss}
\end{equation}
SD incorporates the text conditioning using $\mathbf{t_p}$ into the denoising process via \textit{cross-attention}~\cite{rombach2022high} as:
\vspace{-0.2cm}
\begin{equation}
\label{eq:atten}
\small
\left\{ \begin{matrix}Q = W_Qz_t;~K=W_{K}\mathbf{t_p};~ V=W_{V}\mathbf{t_p} \\\mathtt{Attention}(Q,K,V)=\mathtt{SoftMax}(\frac{QK^T}{\sqrt{d}})\cdot V\end{matrix} \right.
\vspace{-0.2cm}
\end{equation}

\noindent where $W_Q$, $W_K$, and $W_V$ are the learnable projection matrices. $W_K$ and $W_V$ linearly projects the text ecoding $\mathbf{t_p}\in\mathbb{R}^{77\times 768}$ to form the \textit{``Key''} and \textit{``Value''} vectors. Whereas, $W_Q$ projects the intermediate noisy latents to form the \textit{``Query''} maps~\cite{rombach2022high}. The cross-attention map is produced as $\mathtt{SoftMax}(\frac{QK^T}{\sqrt{d}})\cdot V$. Essentially, the cross-attention map indicates the correspondence between the textual prompt and the spatial regions of the image~\cite{rombach2022high}.


\vspace{0.1cm}
\noindent \textbf{T2I-Adapter:}
Moving beyond conventional \textit{textual conditioning}~\cite{rombach2022high}, T2I-Adapter \cite{mou2023t2i} enables a myriad of different \textit{spatial conditioning signals} \cite{koley2024its} (\eg, segmentation masks, scribbles, sketches, key pose, depth maps, colour palates, etc., or their weighted combinations) to guide the T2I image generation process of SD~\cite{rombach2022high}. In practice, T2I-adapter \cite{mou2023t2i} trains a lightweight network (comprising one convolutional and four residual blocks) that extracts deep features from spatial conditioning signals at four different scales. Those extracted \textit{conditioning-features} are then added with intermediate features of SD's UNet decoder at each scale~\cite{mou2023t2i} to influence denoising with the given condition \cite{mou2023t2i}.

\vspace{-2mm}
\section{Problem Definition and Challenges}
\vspace{-1mm}
Given a reference portrait photo $\mathcal{I}_{p}$ depicting a specific identity $p$ and a free-hand abstract sketch $\mathcal{S}$ as the query, we aim to generate a caricature $\mathcal{C}_p^s$, which should retain the identity \cite{deng2019arcface} captured in $\mathcal{I}_{p}$, while reflecting the sketch's ($\mathcal{S}$) influence on its shape.
Notably, $\mathcal{I}_{p}$ may represent an individual \textit{not} encountered previously, and the free-hand sketch $\mathcal{S}$ may depict random or highly \textit{unconstrained} deformations \cite{cao2018carigans} or shape exaggerations \cite{shi2019warpgan}, to be reflected in $\mathcal{C}_p^s$.

Complexity stems from the delicate balance between preserving identity \cite{deng2019arcface} and introducing sketch-guided shape deformations \cite{chu2021learning}. Learning the \textit{unique} identity from a single reference image is non-trivial, given the risk of overfitting \cite{gal2023an, ruiz2022dreambooth} on limited data. Adding to it is the complexity of generating an exaggerated shape in \textit{accordance} with the sketch \cite{koley2024you, bandyopadhyay2024INR}. Addressing these challenges requires a robust model capable of learning and generalising \cite{tewel2023keylocked} from a single reference image while optimising the trade-off between identity preservation and shape exaggeration.


\begin{figure*}[!hbt]
\centering
\includegraphics[width=1.\linewidth]{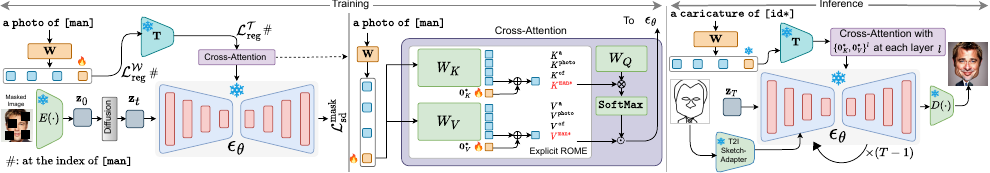}
\vspace{-6mm}
\caption{Within cross-attention layers, Explicit ROME (\cref{sec:rank1}) edits the concept entry with trainable target output $\mathbf{o}^*$ that encapsulates the identity features. We also employ a dynamic masking method (\cref{sec:rmr}), selectively occluding latent regions during training to enhance model robustness. Additional regularisation (\cref{sec:reg}) is applied to word embeddings and text encoding through superclass. During inference, a frozen T2I-sketch-adapter \cite{mou2023t2i} provides shape guidance, resulting in an output caricature with the desired identity and shape. A similar training pipeline is used for the style image as well. We use \cref{eq:style_mixing} to perform sketch+style guided caricature generation.}
\label{fig:architecture}
\vspace{-4mm}
\end{figure*}

\section{Sketch for Caricature Generation}
\vspace{-1mm}
\noindent \textbf{Overview:} Our approach diverges from the prevalent use of pre-trained StyleGAN's \cite{karras2019style, karras2020analyzing, karras2021aliasfree} latent space in facial image editing \cite{patashnik2021styleclip} tasks. Instead, we opt for a pre-trained text-to-image stable diffusion (SD) model \cite{rombach2022high}, known for its generalisation \cite{li2023your} and adaptability \cite{gal2023an} across diverse and wild scenarios. 
Our problem being inherently multi-modal \cite{tang2023emergent}, where $\mathcal{I}_{p}$ is a \textit{real photo}, $\mathcal{S}$ is a black-and-white sparse abstract \cite{hertzmann2020line} \textit{line drawing}, and the output caricature ($\mathcal{C}^s_p$) typically extends \textit{beyond real photo} modality, SD \cite{rombach2022high} becomes an ideal fit as it excels in handling such scenarios which are less encountered \cite{tang2023emergent} in real life.

Our personalised text-to-image (T2I) framework involves fine-tuning the SD model to capture identity in the reference photo $\mathcal{I}_{p}$ and generate the \textit{same} identity in various contexts \cite{gal2023an}. Consequently, we leverage an off-the-shelf T2I-Sketch-Adapter \cite{mou2023t2i} to spatially condition the identity-adapted SD model. This process effectively integrates shape guidance from sketch, aligning $\mathcal{C}^s_p$ with the intended shape.

Our caricature generation pipeline extends further to include style \cite{Jang2021StyleCari} adaptation, by acquiring low-level style features from a single style-reference image $\mathcal{I}g$ characterised by a specific style $g$. The resulting output caricature $\mathcal{C}_{p|g}^s$, now concurrently preserves the identity, style, and shape derived from $\mathcal{I}_p$, $\mathcal{I}_g$, and $\mathcal{S}$, respectively.


\vspace{-1mm}
\subsection{Baseline off-the-shelf Solutions}
\vspace{-1mm}

Given the recent rise of personalised T2I frameworks \cite{saharia2022photorealistic, rombach2022high, ruiz2022dreambooth}, one can naively finetune it using a single reference identity image, and further generate a sketch-conditioned shape-exaggerated output caricature plugging an off-the-shelf T2I-Sketch-Adapter \cite{gal2023an}. Among such frameworks, Textual Inversion \cite{gal2023an} aims to learn a new pseudo word embedding $\mathbf{v}_*$ (representing the concept) in $\mathcal{W}$ space by directly optimising the LDM loss as in \cref{eq:ldmloss} against reference images. Whereas, Perfusion \cite{tewel2023keylocked} further adjusts visual representations through ROME \cite{meng2022locating}, modifying the Value-pathway activation according to the component of all words that are aligned with the target concept.
%
%
%
%


Such a naive solution would however suffer from a few challenges. Firstly, training from a single reference identity ($\mathcal{I}_p$) or style ($\mathcal{I}_g$) image easily leads to overfitting \cite{tewel2023keylocked} in word embeddings, thereby compromising on generalisability to multiple contexts. Secondly, integrating homologous concepts like identity and style, encounters a substantial degree of semantic overlap \cite{tewel2023keylocked}\cut{in the parallel components associated with each concept}. This results in detrimental interference (see \cref{fig:diffusion_2}), causing the concepts to overshadow each other in sketch+style guided caricature generation.
Lastly, being trained on a single reference image only, it fails to generalise towards imbibing the exaggerated \cite{shi2019warpgan} shape guidance from \textit{diverse} sketches \cite{hertzmann2020line}.


Accordingly, we propose three key solutions: \textit{(i)} \emph{Explicit} Rank-1 Model Editing (\cref{sec:rank1}), that edits only at the concept index. Besides preventing potential interference, it also refines the optimisation scope, rendering the adaptation process more effective. Secondly, we implement Random Mask Reconstruction (\cref{sec:rmr}) to enable training with locally masked images, directing the model's focus \textit{away} from local variations and emphasising on key features. This enhances the model's resilience to diverse facial shape constraints \cite{koley2023picture} crucial for caricature synthesis. Thirdly, we incorporate additional regularisation (\cref{sec:reg}) using superclass on word embeddings and text encoding, to counter overfitting, which ensures the model's attention mechanism remains less burdened by the identity \cite{tewel2023keylocked}, allowing more \textit{free}-form shape exaggeration, while preserving identity.

\subsection{Explicit Rank-1 Model Editing}
\label{sec:rank1}

Rank-1 Model Editing (ROME) \cite{meng2022locating} in NLP considers transformer \cite{NIPS2017_3f5ee243} feed-forward layers as memory storage. It utilises learnable outputs to edit this memory, aligning it with the target concept. ROME differentially modifies only the knowledge related to the target, preserving rest of the pre-trained model's memory completely. 
In T2I \cite{mou2023t2i}, textual context is integrated via cross-attention layers, using \emph{`Key'} and \emph{`Value'} pathways akin to feed-forward layers in transformers \cite{tewel2023keylocked}. Our contribution, Explicit Rank-1 Model Editing (Explicit ROME), refines T2I models by applying modifications to the textual encoding locally, specifically \emph{at the position of the concept index} while \emph{leaving} other textual contexts untouched.

Given a reference identity photo $\mathcal{I}_p$ and a textual prompt $p$ = \texttt{`a photo of a P}$^*$\texttt{'}, we convert $p$ to a series of word embedding vectors $p_w$ through word-embedding layer $\mathbf{W}$  where the word embedding corresponding to \emph{concept token} \texttt{P}$^*$ is replaced with a learnable pesudo word embedding vector $\mathbf{v^*}  \in \mathbb{R}^{768}$ for SD v1.5 \cite{rombach2022high}. It is initialised from the word embedding of its corresponding superclass word, like `man' or `woman' based on the gender of the identity photo $\mathcal{I}_p$. Position of the concept token is denoted as $\mathtt{c_i}$. Next, we use a CLIP-text encoder to obtain textual encoding (in $\mathcal{T}$ space) as $\mathbf{t_p} = \mathbf{T}(p_w) \in \mathbb{R}^{77\times 768}$. This $\mathbf{t_p}$ influences the intermediate feature map of SD-UNet through \textit{`Key'} and \textit{`Value'} pathways which we edit via \textit{Explicit Rank-1 Model Editing} in the next stage. 

Considering $W \in \mathbb{R}^{320\times 768}$ (for SD v1.5 \cite{rombach2022high}) from \cref{eq:atten} as the embedding matrix for either \emph{`Key'} as $W_K$ or \emph{`Value'} as $W_V$ and $\mathbf{t_p}$ as the textual encoding, the standard output $h=Wt_p$ is edited by \emph{Explicit ROME} as:

\vspace{-4mm}
\begin{equation}
        h[\mathrm{\mathtt{c_i}}] \leftarrow h[\mathrm{\mathtt{c_i}}] +  s \cdot \Phi(\mathbf{t_p}[\mathrm{\mathtt{c_i}}], i^*) \cdot \ \mathbf{o}^*
   \label{eq:explicit_ROME}
   \vspace{-1mm}
\end{equation}

\noindent where $\Phi(\cdot,\cdot)$ is the cosine similarity function and $\mathbf{o}^*$ is a learnable vector of size $\mathbb{R}^{320}$ (for SD v1.5 \cite{rombach2022high}).  The target input $i^*$ is initialised from CLIP \cite{radford2021learning} text encoding $\mathbf{t_p}$ at $\mathtt{c_i}$ index, and at every step is updated through the exponential moving average \cite{tewel2023keylocked} as $i^* \leftarrow 0.98 \cdot i^*+  \mathbf{t_p}[\mathrm{\mathtt{c_i}}]$. The input $i^*$ serves as a  prototype for gauging the alignment of various contexts with the learned identity. The scale $s$ allows modulating the degree of personalisation during inference, offering more control over results.
The similarity $\Phi(\mathbf{t_p}[\mathrm{\mathtt{c_i}}], i^*)$ represents how closely the input matches $i^*$, after the interaction of learnable pseudo word embedding $\mathbf{v^*}$ and other word embeddings. It can adjust what level of identity features in the caricature should be embedded depending on various contexts (freehand sketch in our case). Instead of complex Mahalanobis distance \cite{tewel2023keylocked} based formulation we utilise the cosine distance to calculate the similarity, which is an intuitive and effective choice according to the text encoder CLIP \cite{radford2021learning}. In all the cross-attention layers' \emph{`Key'} and \emph{`Value'} pathways, we explicitly apply Explicit ROME as in \cref{eq:explicit_ROME} only at the index position of the concept token $\mathtt{c_i}$. This aligns visual features with the target concept, therefore preserving other textual contexts, and consequently ensuring generalisability without compromise. 


Similar to adapting to a reference identity photo  $\mathcal{I}_p$, one can adapt it for a specific style-image $\mathcal{I}_g$ as well, taking superclass word for style images as  `comics', `illustration' etc.  In particular, \cref{eq:explicit_ROME} can be extended to combine multiple independently trained concepts as follows:

\vspace{-3mm}
\begin{equation}
h[\mathrm{\mathtt{c_i}}] \leftarrow h[\mathrm{\mathtt{c_i}}] +  \sum_{j=1}^J s_j \cdot \Phi(\mathbf{t_p}[\mathrm{\mathtt{c_i}}], i^*_j) \cdot \mathbf{o}^*_j
   \label{eq:style_mixing}
   \vspace{-1mm}
\end{equation}

This equation independently treats each concept at its respective $j^\text{th}$ index, preserving unique elements without \textit{unintentional blending}. This ensures easier integration of multiple concepts in caricature synthesis \cite{tewel2023keylocked}, addressing the challenge of blending homologous identity and style \cite{Jang2021StyleCari}.

To sum up, our method has the following trainable parameters: \textit{(i)} a single pseudo word embedding $\mathbf{v^*} \in \mathbb{R}^{768}$. \textit{(ii)} the tuple $\{\mathbf{o_K^*},\mathbf{o_V^*}\}^l$ at each cross-attention layer $l$ for \emph{`Key'} and \emph{`Value'} pathway respectively.
Every $\mathbf{o}^*$ has a dimension of 320, thus making our Explicit ROME overall $30 \times$ lesser learnable parameters than Perfusion \cite{tewel2023keylocked}.

\subsection{Random Mask Reconstruction}
\label{sec:rmr}
One of the major challenges of caricature synthesis is \textit{recreation} of the reference-style \cite{Jang2021StyleCari}, while \textit{maintaining} the subject's unique identity \cite{cao2018carigans, Jang2021StyleCari, tewel2023keylocked, HuoBMVC2018WebCaricature}. To ensure the seamless reproduction of style and identity in the output caricature, we introduce random mask reconstruction (RMR) loss. We hypothesise that random masking of the reference images would shift the model's focus from local spatial regions, enforcing it to understand the global concepts (\textit{i.e.}, style and identity). Given a random masked image, we pass it through the encoder $E(\cdot)$, to obtain a masked latent image $z_0^m$ which after forward diffusion becomes $z_t^m$. This upon passing through UNet-denoiser $\epsilon_\theta$ conditioned on $\mathbf{t_p}$, the modified SD objective becomes: 

\vspace{-0.6cm}
{\begin{equation}
    \mathcal{L}^{\mathrm{mask}}_\text{sd} = \mathbb{E}_{\mathbf{z}_t, t, \mathbf{t_p}, \epsilon}(||(\epsilon-\epsilon_{\theta}(\mathbf{z}_t^m, t, \mathbf{t_p})) \odot M ||^2_2)
    \label{eq:masked_loss}
    \vspace{-0.1cm}
\end{equation}}
where $M$ is the equivalent latent space binary mask with size same as $z_t$. It is used to impose $\mathcal{L}^{\mathrm{mask}}_\text{sd}$ on the unmasked areas only. In practice, we obtain $M$ via bilinear downscaling from a randomly sampled mask \cite{he2021masked} in the pixel space. 

\begin{figure*}[!t]
\centering
   \includegraphics[width=0.98\linewidth]{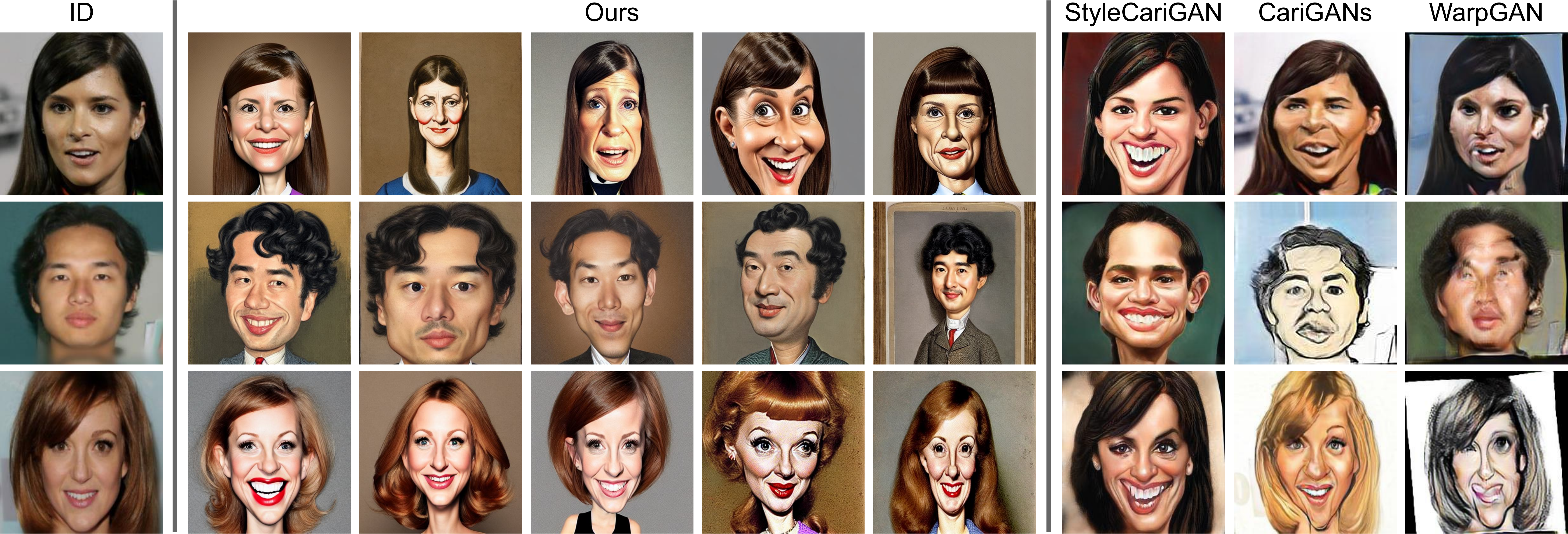}
    \vspace{-0.2cm}
   \caption{\textbf{Qualitative comparison with GAN-based deformation models.} These visual results illustrate our method's higher fidelity and shape flexibility in caricature synthesis compared to existing method \textit{viz.} StyleCariGAN \cite{Jang2021StyleCari}, CariGANs \cite{cao2018carigans}, and WarpGAN \cite{shi2019warpgan}.}
\label{fig:sota}
\vspace{-0.4cm}
\end{figure*}

\subsection{Concept Regularisation}\label{sec:reg}

Any marked deviation of the concept word embedding $\mathbf{v^*}$, risks dominating of the text encoder and attention mechanism by the \textit{concept}, thus losing generalisability to sketch-based deformations. We thus apply $l_2$ regularisation \cite{gal2023an} on the concept word embedding against its superclass word embedding $S_c^w$ in $\mathcal{W}$ space to prevent overfitting of the text encoder. Furthermore, we impose cosine distance-based regularisation loss between text encodings $\mathbf{t_p}$ (using $\mathbf{v^*}$) and $\mathbf{t_p^{s_c}}$ (using $S_c^w$) from CLIP textual encoder $\mathbf{T}$, at the position of concept token $\mathtt{c_i}$. Therefore, the regularisation losses in $\mathcal{W}$ and $\mathcal{T}$ spaces become: 
\vspace{-2mm}
\begin{equation}
\begin{split}
    \mathcal{L}_\text{reg}^{\mathcal{W}} = l_2(\mathbf{v^*}, S_c^w) \;;\;
    \mathcal{L}_\text{reg}^{\mathcal{T}} =  1 - \Phi(\mathbf{t_p}[\mathtt{c_i}], \mathbf{t_p^{s_c}}[\mathtt{c_i}])
\end{split}
\label{eq:regularisation}
\vspace{-0.4cm}
\end{equation}

\noindent Finally, the overall training loss becomes $\mathcal{L}_\text{total} = \mathcal{L}_\text{sd}^{mask} + \lambda_1 \mathcal{L}_\text{reg}^{\mathcal{W}} + \lambda_2 \mathcal{L}_\text{reg}^{\mathcal{T}}$. Please see \cref{fig:architecture} for a summarised overview of training and inference pipelines.

\begin{figure*}[!t]
\centering
   \includegraphics[width=0.97\linewidth, trim={0 15cm 0 0},clip]{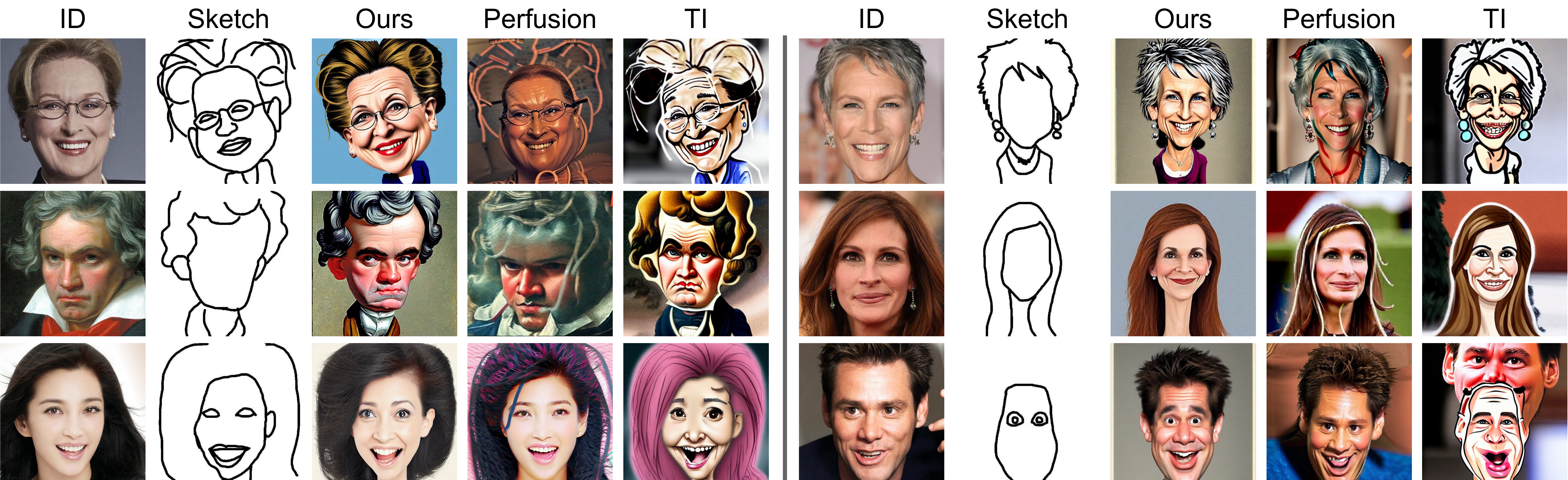}
   \vspace{-0.3cm}
   \caption{\textbf{Comparison with T2I personalisation approaches.} Our framework is stronger in single-image personalisation caricature synthesis against Perfusion \cite{tewel2023keylocked} and TI \cite{gal2023an}.}
   \vspace{-0.3cm}
\label{fig:diffusion_1}
\end{figure*}

\section{Experiments}
\label{sec:experiments}
\textbf{Datasets.}
We use the WebCaricature dataset \cite{HuoBMVC2018WebCaricature} to source identities and styles. To validate our approach via a quantitative comparison and a user study, we curate a test dataset encompassing 20 identities, 4 styles, and 12 distinctive edge maps as shapes. These edge maps are extracted from caricature images of WebCaricature  \cite{HuoBMVC2018WebCaricature}, leading to 960 unique caricature pairs for evaluation. For a fair assessment of our method, the carefully selected identities encompass a wide-spectrum of race, gender, and age, thereby upholding diversity and inclusiveness in our evaluation. Analysing qualitative results, we incorporate amateur freehand sketches, incorporating real user interpretation into the assessment. 

\vspace{0.1cm}
\noindent\textbf{Implementation Details.}
Our implementation is based on Stable Diffusion v1.5 \cite{rombach2022high}. We train using AdamW \cite{loshchilov2018decoupled} optimiser, with a batch size of $16$, learning rates $0.2$ and $0.002$ for target outputs and embeddings respectively.
Fine-tuning consists of $40$ and $100$ steps for identities and styles, respectively. We conduct all experiments on a single NVIDIA GTX $4090$ GPU, taking $1$ minute for identity and $2$ minutes for style fine-tuning. For inference, results are sampled with 50 steps along with a classifier-free guidance \cite{ho2021classifierfree} scale of $9$. We use the prompts $\mathtt{``a~caricature~of~[id*]"}$ and $\mathtt{``a~caricature~of~[id*]~in~the~style~of~[style*]"}$ to generate caricatures.

\begin{figure*}[!htbp]
\centering
   \includegraphics[width=0.97
   \linewidth]{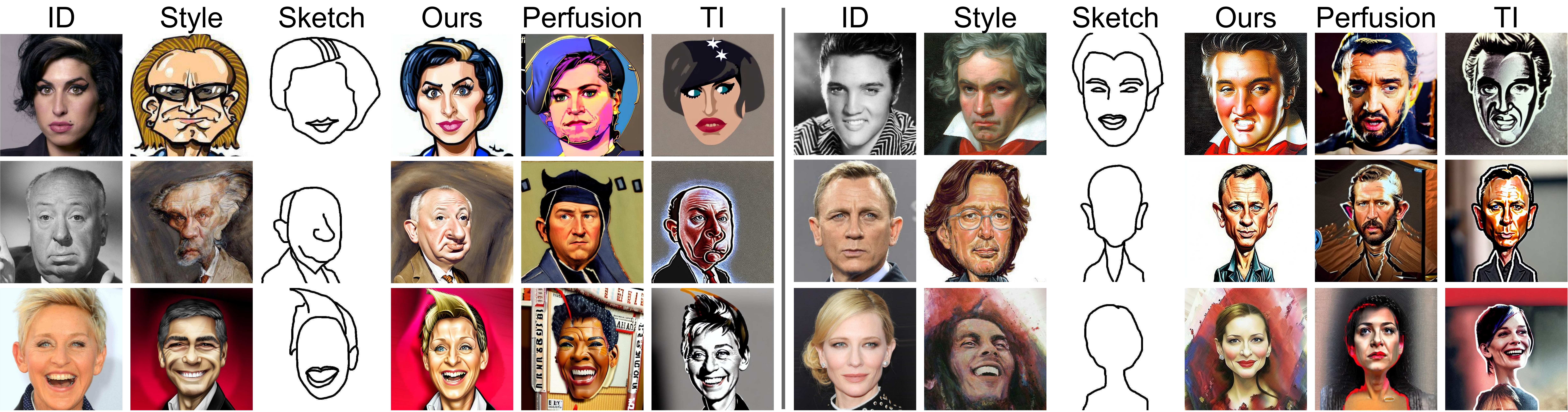}
   \vspace{-0.3cm}
   \caption{\textbf{Comparison with T2I personalisation approaches with style reference.} Demonstrates our model's robustness in generating stylised caricatures with faithful identity and style, surpassing other methods like Perfusion \cite{tewel2023keylocked} and TI \cite{gal2023an}.}
   \vspace{-0.4cm}
\label{fig:diffusion_2}
\end{figure*}

\vspace{-0.1cm}
\subsection{Qualitative Evaluation}
\cref{fig:teaser} shows the efficacy of our proposed method in generating caricatures while faithfully conforming to specifications of identity \cite{HuoBMVC2018WebCaricature}, style \cite{Jang2021StyleCari}, and sketch shape \cite{hertzmann2020line}. Given a subject, our method demonstrates its robust caricature synthesis \cite{Jang2021StyleCari} potential in the above half of \cref{fig:teaser}. It moves beyond the traditional confines of feature scaling \cite{shi2019warpgan, cao2018carigans, Jang2021StyleCari} to a paradigm where features can be adjusted and exaggerated with ease of using sketch-based guidance.
Such flexibility reaches into the domain of fine-grained facial feature manipulation \cite{chenDeepFaceDrawing2020} adjusting shape, features (mouth, ears, nose), expressions, as well as hairstyles, while also attending to accessories and novel perspectives.
From simple one-stroke outlines to intricate details, our model demonstrates adaptability to varying sketch complexities. Remarkably, it achieves this without reliance on identity-tailored components \cite{ruiz2023hyperdreambooth}, capturing the subtle essence of human faces from merely a \textit{single} reference image with only a \textit{few} fine-tuning steps. Our framework addresses the challenge of identity preservation while applying exaggeration and distortion, exemplifying a robust resistance to overfitting. When constrained by a sketch, the model seamlessly integrates identity into the shape, ensuring recognisability without apparent visual artefacts, while maintaining the prior knowledge of the SD \cite{rombach2022high} model.

Lower half of \cref{fig:teaser} illustrates our model's ability to harmonise two conflicting concepts: identity and style, each derived from separate human likenesses. The objective is to unify them within a single synthesised caricature face. Diffusion backbones \cite{rombach2022high} usually struggle with such duality, yet our model overcomes this, rendering caricatures with high fidelity to both identity and style elements. 

\subsection{Comparison with SOTA}
\vspace{-0.1cm}
\label{subsec:comparison}
We benchmark our caricature synthesis against three state-of-the-art (SOTA) deformation-based models \textit{viz.} \textbf{StyleCariGAN}~\cite{Jang2021StyleCari}, \textbf{CariGANs}~\cite{cao2018carigans}, and \textbf{WarpGAN}~\cite{shi2019warpgan}. These models however do not support caricature synthesis with combined conditioning on identity \cite{cao2018carigans}, style \cite{Jang2021StyleCari}, and shape \cite{HuoBMVC2018WebCaricature} like ours. We extend our comparison to advanced SD-based~\cite{rombach2022high} personalisation models, like \textbf{Textual Inversion (TI)} \cite{gal2023an} and \textbf{Perfusion} \cite{tewel2023keylocked} as well.

\textbf{CariGANs} \cite{cao2018carigans} and \textbf{WarpGAN} \cite{shi2019warpgan} which rely on landmarks and control point manipulation, clearly show distortions and artefacts in \cref{fig:sota}.
By leveraging deep feature-map modulation from StyleGAN \cite{karras2019style}, \textbf{StyleCariGAN} \cite{Jang2021StyleCari} delivers higher-fidelity caricatures, yet it is limited to {pre-defined scale-based exaggeration} \cite{Jang2021StyleCari}, ignoring shape information.
%
On the other hand \textbf{TI} \cite{gal2023an} and \textbf{Perfusion} \cite{tewel2023keylocked} fail to preserve identity in caricatures due to overfitting caused by single-image personalisation (\cref{fig:diffusion_1}).
%
Furthermore, lacking an effective interaction-control mechanism, they suffer from (\cref{fig:diffusion_2}) identity and style ambiguity, thus deviating from corresponding references. 
%
%
Our Explicit ROME strategy circumvents these pitfalls, ensuring targeted editing at corresponding positions without disrupting other text and concept encodings in the cross-attention mechanism~\cite{mou2023t2i}, as verified by our superior qualitative results.

\begin{figure*}[t]
\centering   \includegraphics[width=1.\linewidth]{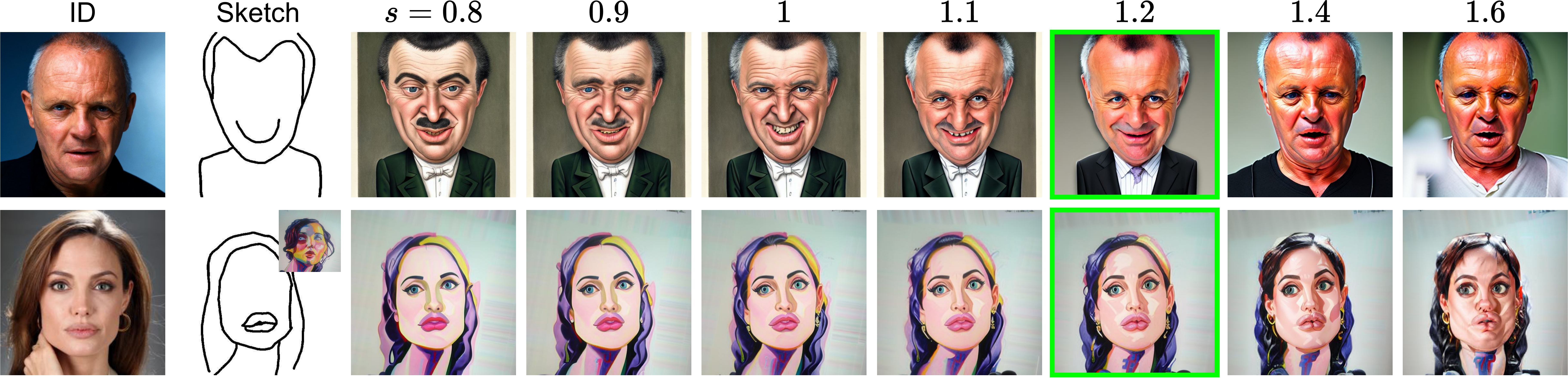}
    \vspace{-0.6cm}
   \caption{\textbf{Identity Scale Adaptability.} Our method provides a dynamic adjustment of the identity scale $s$, exemplifying flexibility.}
\label{fig:scales}
\vspace{-0.3cm}
\end{figure*}

\begin{figure}[!htbp]
\centering
\includegraphics[width=1\linewidth,trim={0 3cm 0 0}]{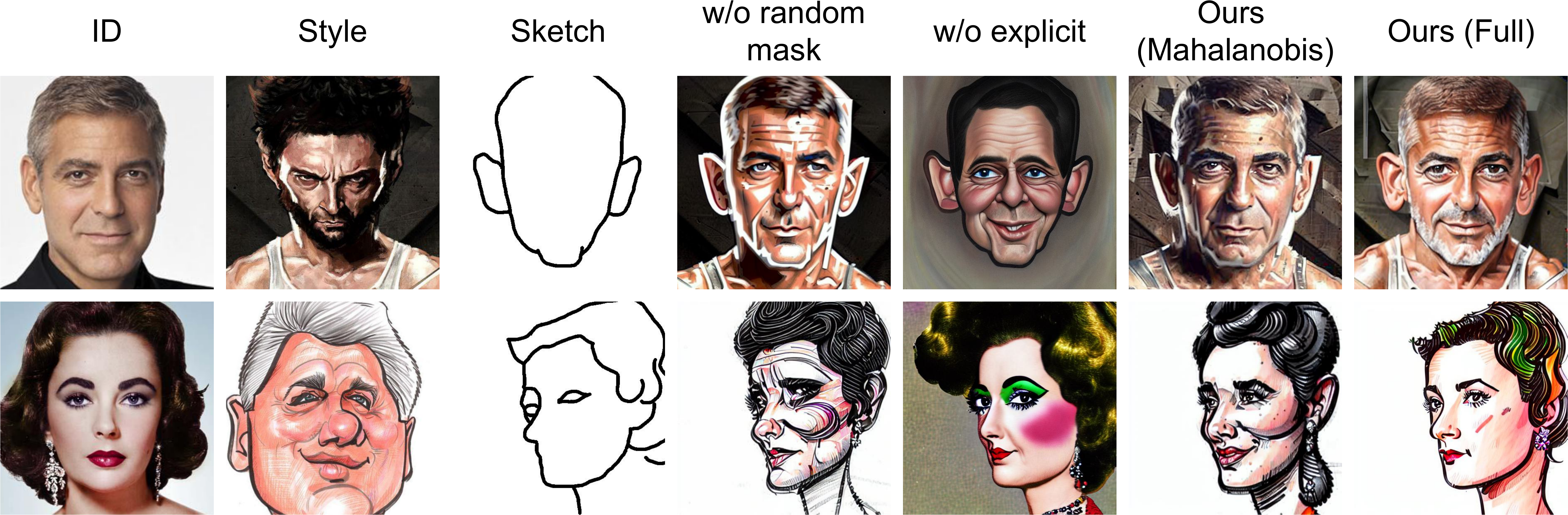}
\vspace{-4mm}
   \caption{Qualitative results of our ablation study}
\label{fig:ablation}
\vspace{-2mm}
\end{figure}

Now, for quantitative evaluation (\cref{tab:quant}) , we use CLIP-Score~\cite{radford2021learning} on \textit{ID}, \textit{Style}, and \textit{Shape}. It measures \textit{ID}/\textit{style}-fidelity as the similarity between generated caricatures and ID/style images using a pre-trained CLIP \cite{radford2021learning} encoder, and \textit{shape} fidelity as the same between edgemaps of generated caricatures and conditioning sketches. Notably, at the same level of shape similarity, our results have the highest identity and style similarity at $0.671$ and $0.576$, which is $3\times$ and $10\times$ faster than \textbf{Perfusion} \cite{tewel2023keylocked} and \textbf{TI} \cite{gal2023an} respectively. Notably, this was achieved within three minutes of fine-tuning for identity and style.

\vspace{-2mm}
\begin{table}[t]
\setlength{\tabcolsep}{10pt}
\renewcommand{\arraystretch}{0.9}
\centering
\footnotesize
\caption{\textbf{Quantitative comparison.} Quantitative metrics of various approaches and our framework ablative design, reflecting the precise quantitative edge our model holds over existing methods.}
\vspace{-1mm}
\vspace{-0.2cm}
    \begin{tabular}{lccc}
        \toprule
        Methods               & ID $\uparrow$   & Style $\uparrow$ &  Shape $\uparrow$ \\
        \cmidrule(lr){1-4}
        TI~\cite{gal2023an}                  & 0.634 & 0.553 & 0.633 \\
        Perfusion~\cite{tewel2023keylocked}  & 0.536 & 0.549 & 0.676 \\ \cmidrule(lr){1-4}
        Ours (w/o rand mask)                 & 0.659 & 0.567 & 0.694 \\
        Ours (w/o explicit)                  & 0.664 & 0.530 & 0.661 \\
        Ours (Mahalanobis)                   & 0.666 & 0.574 & 0.663 \\
        \rowcolor{gray}
        \cellcolor{YellowGreen!40}\textbf{Ours-full} & \cellcolor{YellowGreen!40}0.671     & \cellcolor{YellowGreen!40}0.576 & \cellcolor{YellowGreen!40}0.654 \\
        
        \bottomrule
    \end{tabular}
\vspace{-6mm}
\label{tab:quant}
\end{table}

\vspace{0.1cm}
\noindent\textbf{Human Study.} We conduct a thorough human study to judge the efficacy of our method from end-users' perspective. Specifically, each of the $15$ users were shown $20$ tuples, each containing $\{$ID, input sketch, style image, output caricature$\}$ from \textit{all} competing methods, and asked to rate the caricatures on a discrete scale of $[1,5]$ (\textit{worst} to \textit{best}) based on fidelity to input \textit{sketch-shape}, \textit{style}, and \textit{ID} --  resulting in a total of $300$ responses per method.
The final score for each method is calculated from the mean of all its responses. Our method with high shape-fidelity and identity-preservation, garners an impressive overall score of $4.1$ (\cref{tab:user_study}) surpassing others. Although, the users preferred \textbf{TI} \cite{gal2023an} over \textbf{Perfusion} \cite{tewel2023keylocked} in terms of identity-preservation, they both score lower compared to ours.

\begin{figure}[t]
\centering
   \includegraphics[width=1.\linewidth]{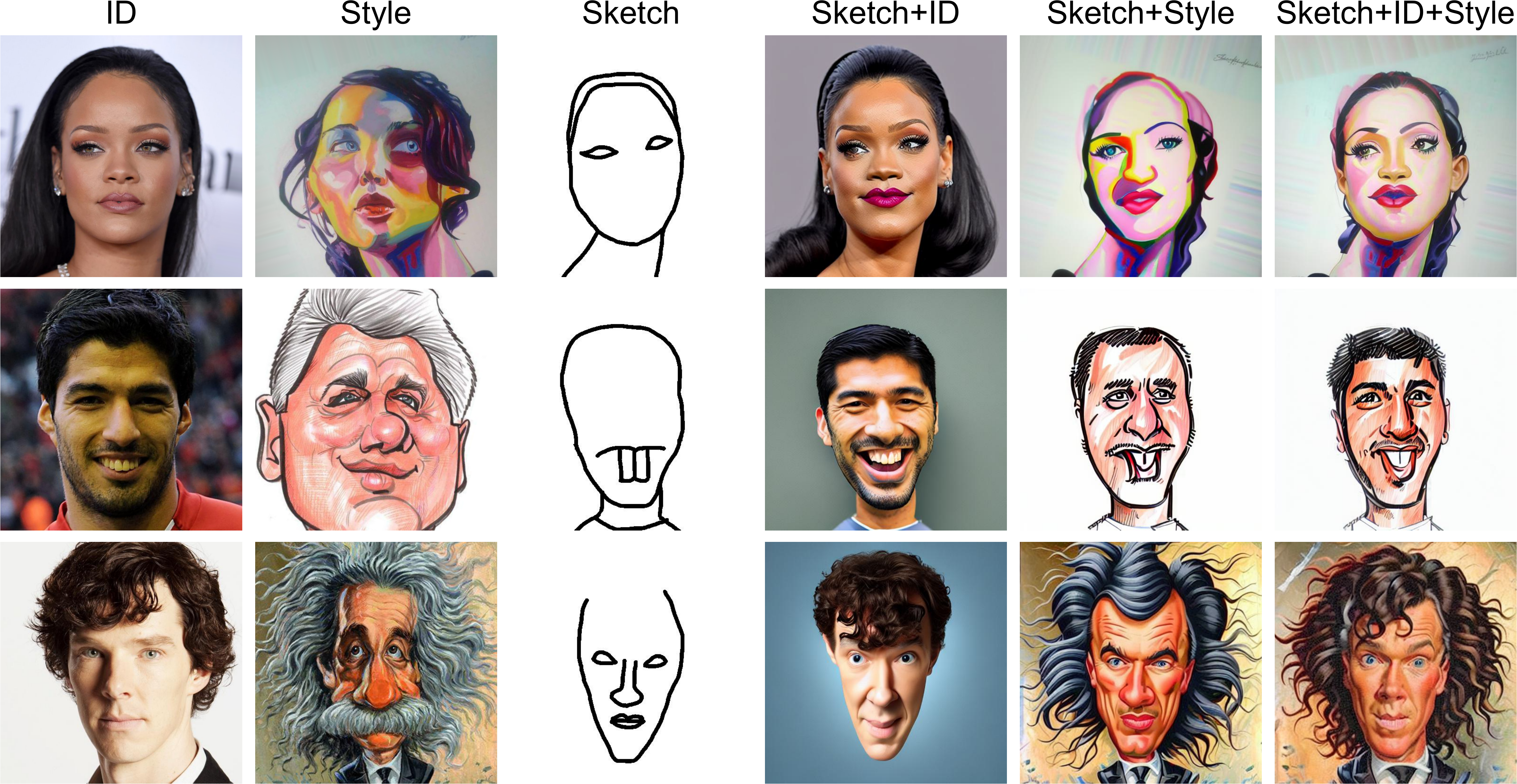}
   \vspace{-0.6cm}
   \caption{Our model's capacity to integrate various modalities.}
   \vspace{-0.6cm}
\label{fig:ablation_mod}
\end{figure}

\vspace{-2mm}
\begin{table}[!htbp]
\setlength{\tabcolsep}{8pt}
\renewcommand{\arraystretch}{0.9}
\centering
\footnotesize
\caption{\textbf{Human Study Scores.}}
\vspace{-0.3cm}
    \begin{tabular}{lcccc}
        \toprule
        Methods               & ID $\uparrow$   & Style $\uparrow$ & Shape $\uparrow$ & Overall $\uparrow$ \\
         \cmidrule(lr){1-5}
        TI~\cite{gal2023an}                  & 3.3 & 2.5 & 2.9 & 2.9 \\
        Perfusion~\cite{tewel2023keylocked}  & 2.8 & 3.1 & 2.4 & 2.7 \\
        \cellcolor{YellowGreen!40}\textbf{Ours}                 & \cellcolor{YellowGreen!40}4.4 & \cellcolor{YellowGreen!40}3.8 & \cellcolor{YellowGreen!40}4.2 & \cellcolor{YellowGreen!40}4.1 \\
        
        \bottomrule
    \end{tabular}
\label{tab:user_study}
\vspace{-4mm}
\end{table}

\subsection{Ablation Study}
\label{subsec:ablation}
\vspace{-1mm}

\textbf{Design Choices.} Our ablation experiments are depicted in \cref{fig:ablation} and \cref{tab:quant}.
\textit{(i)} To judge the impact of explicit editing we exclude it for an experiment, to observe that caricatures lose defining visual characteristics, dropping scores to $0.006$ and $0.046$ in identity and style similarities, respectively, thus proving its significance.
\textit{(ii)} Removing Random mask reconstruction (\cref{sec:rmr}) results in $0.659$ ($0.553$) for ID (style), validating its role in reinforcing robustness against local distortions in personalisation \cite{tewel2023keylocked}. 
\textit{(iii)} The replacement from the Mahalanobis distance to applying cosine similarity on the Euclidean distance alleviates the need for cumbersome pre-cached uncentered covariance estimation \cite{tewel2023keylocked}, leading to a more streamlined training process. More importantly, it causes an apparent improvement in visual quality, and a slight increase ($0.05/0.002$ in ID/style) in the similarities as well, thus replacing cosine similarity with a more efficient choice.

\vspace{0.1cm}
\noindent\textbf{Modalities.}
While the fourth column of \cref{fig:ablation_mod} validates our model's precision in preserving identity, the fifth displays our integration of styles with shapes. Finally, the sixth column highlights our Sketch+ID+Style result, achieving high fidelity to input ID \cite{HuoBMVC2018WebCaricature}, sketch \cite{hertzmann2020line} and style \cite{Jang2021StyleCari}.

\vspace{0.1cm}
\noindent\textbf{Impact of Identity Scale ($s$).}
\cref{fig:scales} shows the influence of identity scale $s$ on the generated caricatures. Evidently, a higher $s$ tends to \textit{retain} a higher proportion of identity traits in a caricature and vice-versa. Around the sweet spot below $1.40$, users can freely choose this balance as per their own subjective tastes to obtain coherent personalised caricatures \cite{tewel2023keylocked}. In all our experiments we had set $s$ as $1.2$, empirically.

\vspace{-0.2cm}
\section{Conclusion}
\label{sec:conclusion}
\vspace{-0.1cm}
In conclusion, our work marks a significant leap in democratising caricature generation, offering individuals an effortless means to craft personalised artworks with minimal input -- just a photo and a conceptual sketch. By navigating the delicate balance between abstraction and identity, our proposed Explicit Rank-1 Model Editing and Random Mask Reconstruction, empower users to seamlessly merge their unique identity and desired artistic style in the caricature synthesis process. We emphasise that our intention is not to replace the irreplaceable touch of artists but to remove accessibility barriers, allowing enthusiasts to engage in the creative realm of caricature art. More generally, our contribution underscores the potential for AI to harmoniously collaborate with human creativity, ensuring that art remains a captivating and inclusive expression for all.


{
    \small
    \bibliographystyle{ieeenat_fullname}
    \bibliography{main}
}

\clearpage
\setcounter{section}{0}
\renewcommand{\thesection}{\Alph{section}}


\onecolumn{
\begin{center}
    \Large{Supplementary Material for\\\textbf{DemoCaricature: Democratising Caricature Generation with a Rough Sketch}}
\end{center}
}

\section{On conflict between ID and Sketch}

\begin{figure}[h]
    \centering
    \vspace{-6mm}
    \includegraphics[width=0.6\linewidth]{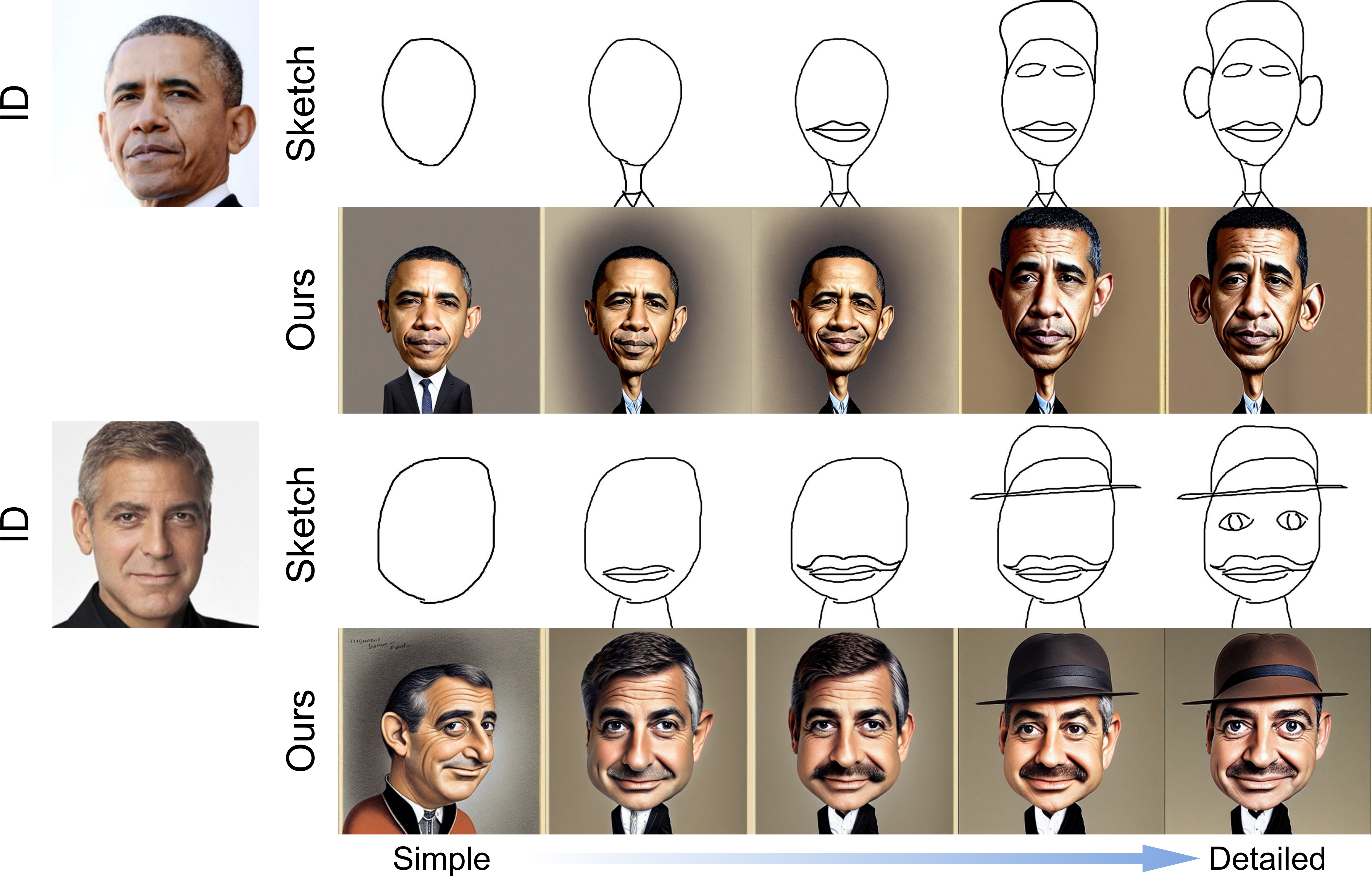}\\
    \vspace{2mm}
    \includegraphics[width=0.6\linewidth]{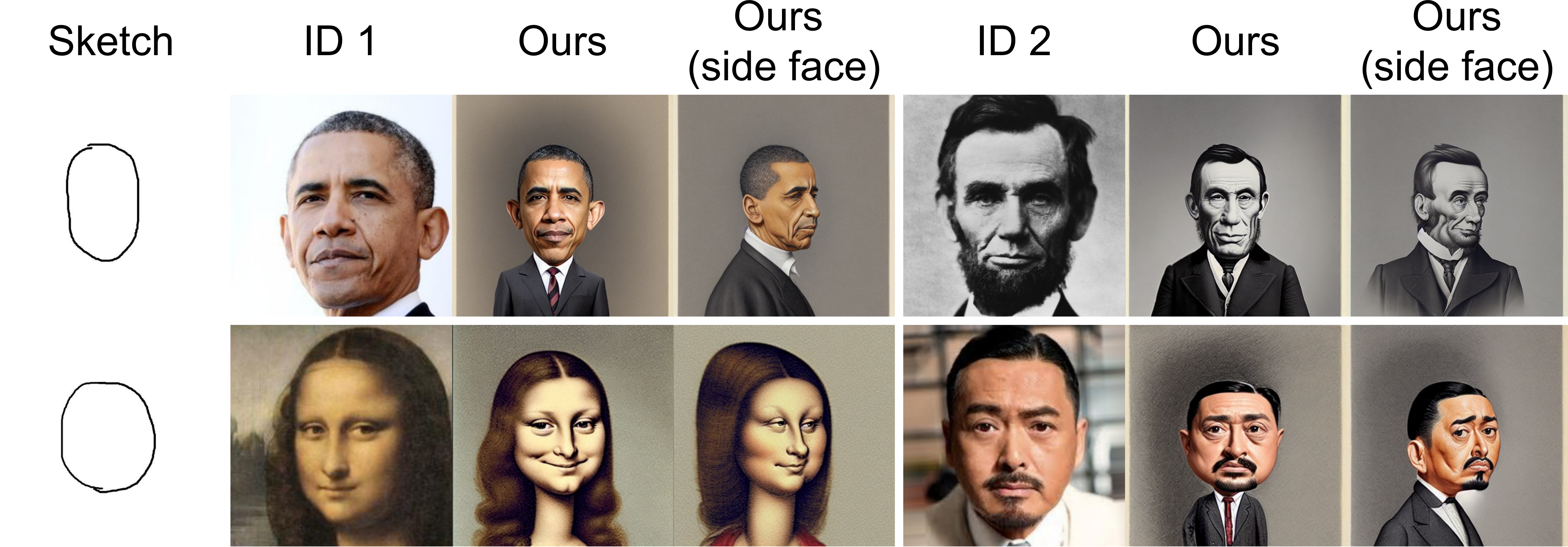}
    \caption{\textit{Top:} Abstraction level. \textit{Bottom:} View angle.}
    \label{fig:rebuttal}
    \vspace{-5.5mm}
\end{figure}

\cref{fig:rebuttal} (top-half), shows a gradual generation process starting from a contour. As simple sketches provide weaker constraints, the model finds it relatively easier to integrate ID and shape. Contrarily for complex sketches, conflict may rise if the sketch is \textit{inconsistent} with the \textit{character} itself. Even under such scenarios our method offers decent results, thus highlighting our method's calibre at \textit{finding a good balance} amidst this conflict. Furthermore, Fig.\ \red{6}, shows how our model balances by adapting the ID scale. While a stronger ID scale can align the results more with the single ID reference, a slightly smaller one can make the results more flexible while \textit{maintaining distinctive characteristics} of the ID. In fact, our method can synthesise completely different perspectives under imprecise sketch shapes (\cref{fig:rebuttal} bottom-half). Such generality is pivotal to solving conflict. 

\section{Details on Random Mask Reconstruction}
\begin{figure}[!htbp]
\centering
   \includegraphics[width=.6\linewidth]{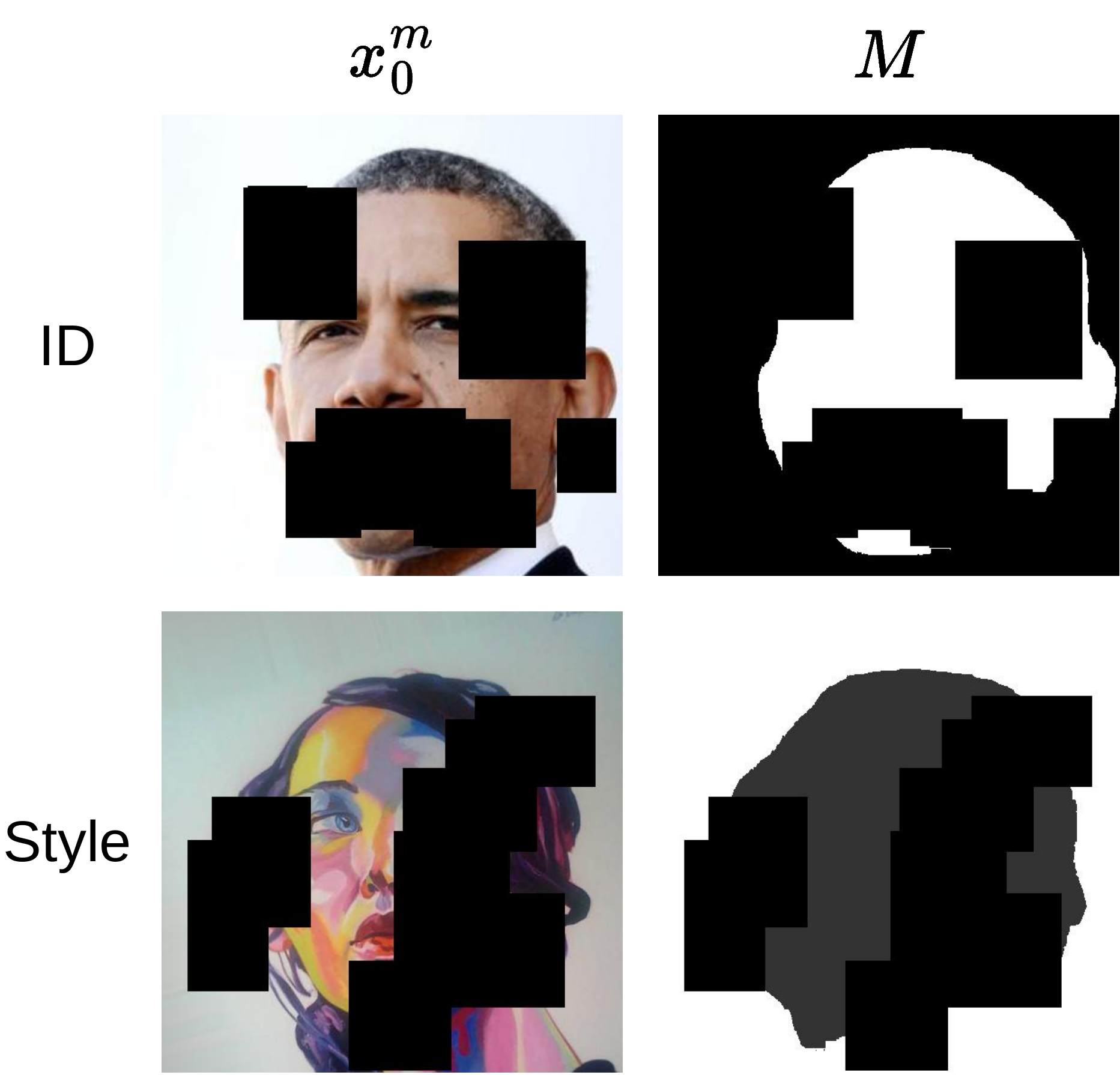}
   \caption{\textbf{Random Mask Reconstruction.} $x_0^m$ mimics an image with local variation, a critical feature of caricature. $M$ makes the objective function focus on the region of interest and ignore the occluded area.}
\label{fig:rmr}
\end{figure}

We present more implementation details on the random mask reconstruction (RMR) as shown in \cref{fig:rmr}.
We create the masked image $x_0^m$ by randomly occluding several patches with different ratios to simulate a caricature having local variation.
Besides the random occlusion, we also apply id- and style-specific masks on $M$ to isolate regions of interest when calculating loss using \cref{eq:masked_loss}.
Specifically, for identity finetuning, $M$ contains a binary mask over the background, making the model capture the distinguishing facial features exclusively.
For style reference, $M$ adopts a small value (0.2 in this work) on the face area, subtly nudging the model to infuse the stylistic elements from both the background and the face into caricatures.

\section{More Qualitative Results}
\begin{figure*}[!htbp]
\centering
   \includegraphics[width=.95
   \linewidth]{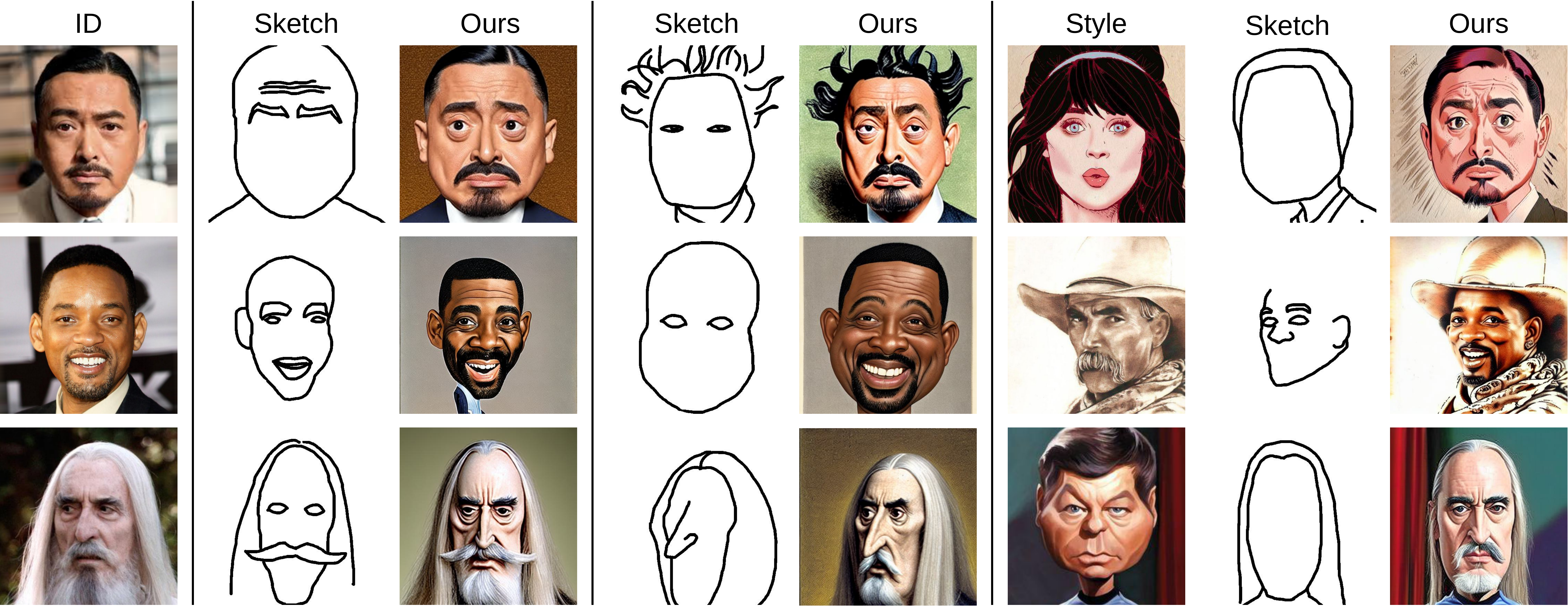}
   \caption{Results on celebrities.}
   \vspace{-0.4cm}
\label{fig:suppl_celeb}
\end{figure*}

\begin{figure*}[!htbp]
\centering
   \includegraphics[width=.95
   \linewidth]{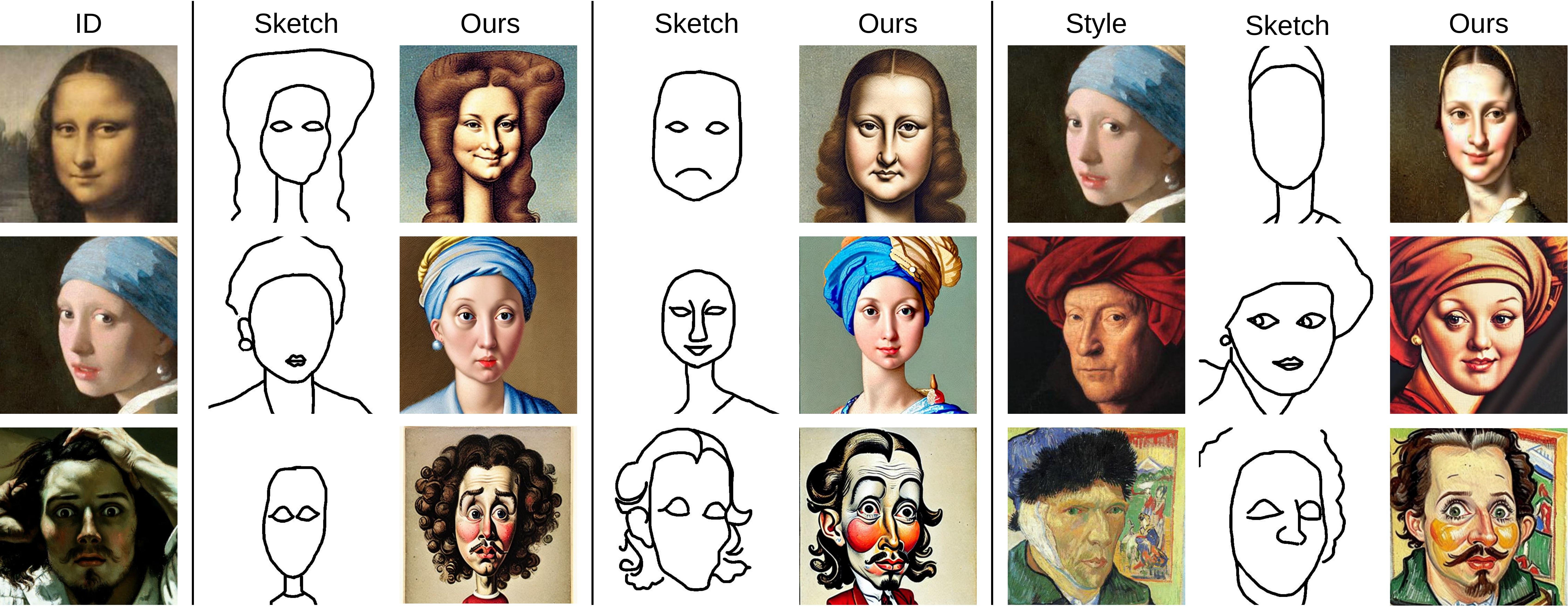}
   \caption{Results on famous artworks.}
   \vspace{-0.4cm}
\label{fig:suppl_art}
\end{figure*}

\begin{figure*}[!htbp]
\centering
   \includegraphics[width=.95
   \linewidth]{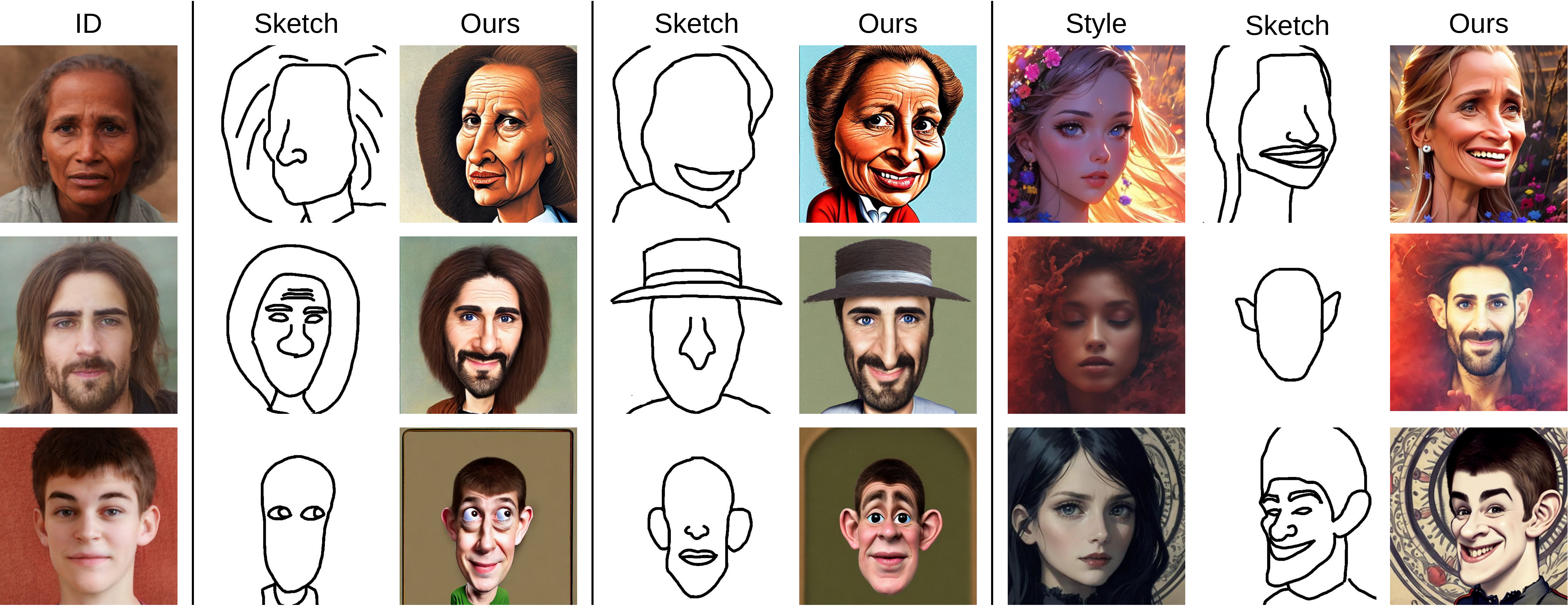}
   \caption{Results on synthetic human faces.}
   \vspace{-0.4cm}
\label{fig:suppl_synthesis}
\end{figure*}

Finally, we provide additional qualitative results across various fields. \cref{fig:suppl_celeb} showcases more caricatures of celebrities.
Beyond celebrities, we exhibit our method's capability of learning identity and style from artistic and synthetic \footnote{We collect the synthetic id image from Ruiz \etal \cite{ruiz2023hyperdreambooth} and synthetic style reference from https://civitai.com.} portraits in \cref{fig:suppl_art} and \cref{fig:suppl_synthesis} respectively.
The results demonstrate the versatility of our democratising caricature generation, allowing users to flexibly and artistically create caricatures with the desired identities and styles.


\end{document}